\documentclass[conference, a4paper]{IEEEtran}

\IEEEoverridecommandlockouts

	\usepackage[turkish]{babel}
	\usepackage[utf8]{inputenc} 
	\usepackage[T1]{fontenc}

\usepackage{multirow}
\usepackage{cite}
\usepackage{xcolor}
\usepackage{amssymb}

\ifCLASSINFOpdf
  \usepackage[pdftex]{graphicx}
\else
\fi
\usepackage[cmex10]{amsmath}

\usepackage{url}
\usepackage{multirow}
\usepackage{array}
\usepackage[lofdepth,lotdepth]{subfig}

\AtBeginDocument{%
  \renewcommand\tablename{TABLE}
}

\AtBeginDocument{%
  \renewcommand\figurename{Figure}
}

\AtBeginDocument{%
  \renewcommand\refname{References}
}

\AtBeginDocument{%
  \renewcommand\abstractname{Abstract}
}

\begin{document}


%

\title{Improving the Explain-Any-Concept by Introducing  \\ Nonlinearity to the Trainable Surrogate Model}



\author{
\IEEEauthorblockN{Mounes Zaval}
\IEEEauthorblockA{Ozyegin University \\ Istanbul, Turkey\\
mounes.zaval@ozu.edu.tr}

\and
\IEEEauthorblockN{Sedat Ozer}
\IEEEauthorblockA{Ozyegin University \\ Istanbul, Turkey\\
sedat.ozer@ozyegin.edu.tr} 
}


%

\maketitle

\begin{abstract}\footnote{This work is accepted for oral presentation at IEEE, 32nd Signal Processing and Communications Applications Conference (SIU), 2024. }
In the evolving field of Explainable AI (XAI), interpreting the decisions of deep neural networks (DNNs) in computer vision tasks is an important process. While pixel-based XAI methods focus on identifying significant pixels, existing concept-based XAI methods use pre-defined or human-annotated concepts. The recently proposed Segment Anything Model (SAM) achieved a significant step forward to prepare automatic concept sets via comprehensive instance segmentation. Building upon this, the {\it Explain Any Concept} (EAC) model emerged as a flexible method for explaining DNN decisions. EAC model is based on using a surrogate model which has one trainable {\it linear} layer to  simulate the target model. In this paper, by introducing an additional nonlinear layer to the original surrogate model, we show that we can improve the performance of the EAC model. We compare our proposed approach to the original EAC model and report improvements obtained on both ImageNet and MS COCO datasets. 
\end{abstract}


\IEEEpeerreviewmaketitle

\IEEEpubidadjcol

\section{Introduction}
In the rapidly evolving domain of Deep Neural Networks (DNNs) based computer vision applications such as image classification \cite{b1}, object detection \cite{sahin2022yolodrone,b2}, autonomous systems \cite{valiente2020connected}, semantic or instance segmentation \cite{OZER2}, and infrared image analysis \cite{OZER}, the interpretability of DNNs is a significant concern. This is primarily due to the {\it black-box} nature of DNNs. Explainable AI (XAI) emerges as a crucial field to find methods that are explainable, and aims to make AI systems more transparent and understandable for humans \cite{b4}. 
In the expanding domain of this field, interpretations vary in quality and effectiveness, leading to inherent compromises. To measure the interpretation, we rely on three principal criteria: faithfulness, understandability, and efficiency \cite{b5, b6, b7}. Faithfulness measures how closely an explanation mirrors the DNN's internal decision-making stage. Understandability gauges whether humans can easily comprehend the explanations. Efficiency considers the computational resources required to generate these explanations.

Unlike pixel-based explanation methods as in \cite{b8, b9}, concept-based explanations utilize higher-level features or 'concepts' within images, such as identifiable objects or patterns, with the aim of making explanations more comprehensible. The Segment Anything Model (SAM) \cite{sam}, a significant recent innovation in this area, enabled automatic generation of concept sets from given images, enhancing the scope of concept-based explanations.

\begin{figure*}[t]
	\centering
	\shorthandoff{=}  
	\includegraphics[scale=0.7]{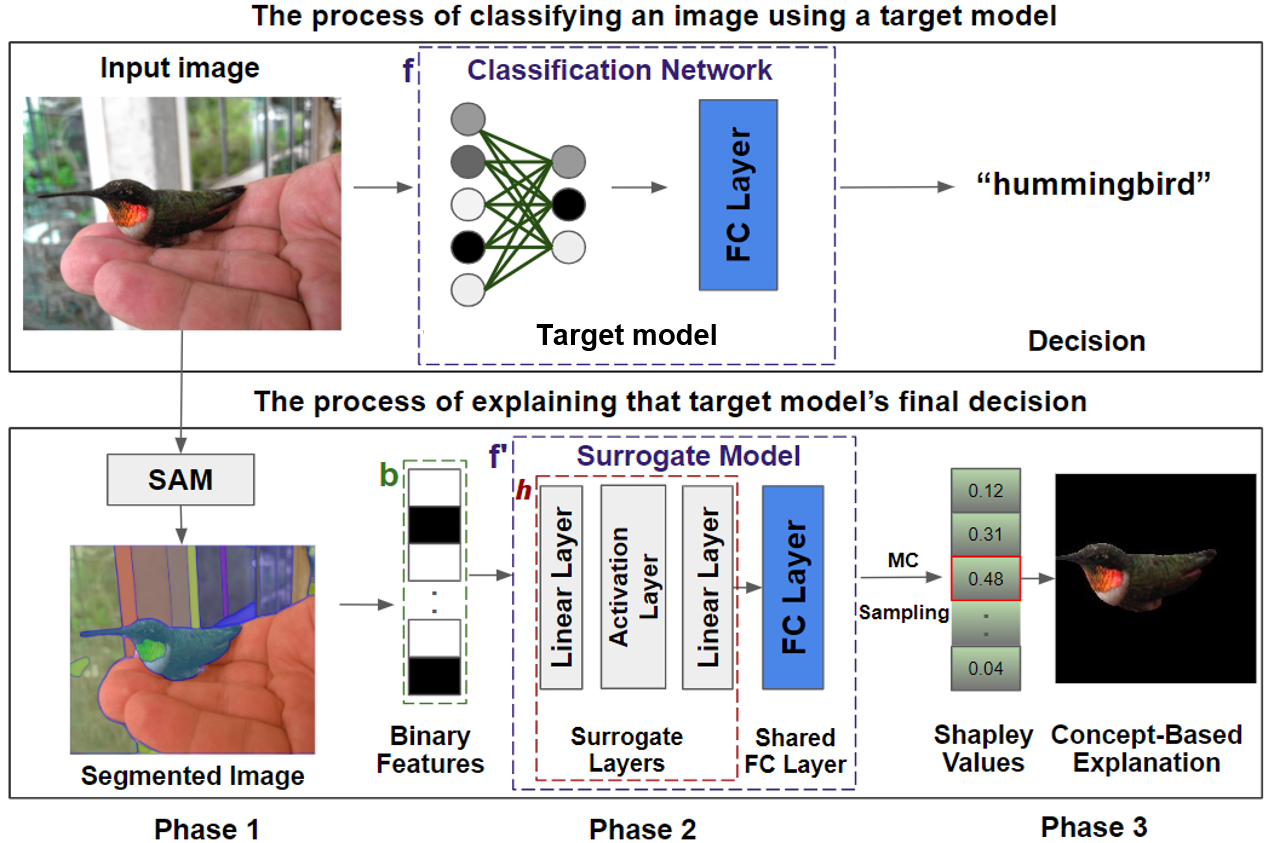}
	\shorthandon{=} 
	\caption{An overview of  the explainability process based on the Explain Any Concept (EAC) \cite{eac}. The top figure shows applying a given classification network on an image and its output. The bottom figure shows the stages of the explainability process including Segment Anything Model (SAM) \cite{sam}, the surrogate model (our version) and the Shapley values.  }
	\label{dem}
\end{figure*}

Recently, Explain Any Concept (EAC) \cite{eac} model is proposed which uses SAM as its backbone. The EAC architecture uses binary features as its inputs and those binary features are then fed into a surrogate model which is connected to the shared fully connected (FC) layer then Shapley values \cite{shap} are calculated. The shared FC layer is the same last FC layer (the output layer) of the original (target) network. Therefore, there is only one linear layer (the surrogate layer) that is trained in the EAC model \cite{eac}. That surrogate layer acts as the feature extractor. However, referencing the universal approximation theorem, we think that using only a single linear layer may not well model the feature extraction properties of the original deep architecture and this issue was also discussed in \cite{eac}. Consequently, in this paper, we study to see if using a nonlinear layer in front of the original layer, benefits the explainability. Therefore, while Shapley value-based methods in XAI frequently used, they involve complex calculations, often necessitating using approximating methods including Monte Carlo sampling \cite{mc}, model-specific approximations \cite{modelspec1,modelspec2}, or (smaller) surrogate models \cite{surrogate1,surrogate2}. These approaches, however, can suffer from low faithfulness due to differences from the target model. This paper diverges from the pursuit of computational gains, instead introduces architectural improvements in the surrogate model to potentially enhance the quality and fidelity of the final explanations, while introducing as minimal number of computations as possible, additionally. Our approach aims to look for an answer whether such architectural modifications in the surrogate model can lead to more accurate and meaningful explanations in the EAC framework.\\
Our contributions in this paper include (i) introducing nonlinearity for the surrogate model in the EAC framework;
(ii) introducing further explainability for the EAC framework in terms of the area under the curve (AUC) value as demonstrated by our experimental results where our Enhanced EAC (EEAC) approach yielded better AUC values for both ImageNet and COCO datasets, when compared to the original EAC framework.

\section{Related Work}

\textbf{XAI Methodologies:} Among the DNN-based techniques, the field can be broadly categorized into two: (i) backpropagation-based and (ii) perturbation-based techniques. (i) Backpropagation-based (or gradient-based methods), aim to determine the influence of different attributes of the input on the final model decisions as in saliency maps\cite{silency}, DeepLIFT\cite{deeplift} and Grad-CAM\cite{b9}. (ii) Perturbation-based models, on the other hand, introduce perturbations on the input and observe changes on the output, sample methods are Local Interpretable Model-agnostic Explanations (LIME) \cite{lime} and SHAP SHapley Additive exPlanations (SHAP) \cite{shap2}.

\textbf{Pixel-Based vs. Concept-Based XAI:} Pixel-based methodologies focus on techniques at the individual pixel level, often employing techniques like saliency maps and attention mechanisms. Though insightful, they are limited by their dependency on model architecture. Concept-based XAI, utilizes Concept Activation Vectors (CAVs) \cite{cav}. It distinguishes between random images and the images containing user-specified concepts. By doing so, it quantifies the influence of concepts on the output. This method, while offering enhanced interpretability, is constrained by the need for predefined human-specified concepts.\\
The EAC pipeline \cite{eac} is a concept-based explanation method, where it utilizes the Segment Anything Model \cite{sam} for automatically highlighting important concepts. The Segment Anything Model (SAM) is designed to produce high-quality object masks from input prompts such as points or boxes. It is capable of generating masks for a large set of objects in an image and has been trained on a very large dataset. SAM demonstrates strong zero-shot performance across various segmentation tasks. 

\section{Enhanced EAC framework}

\begin{figure*}[t]
	\centering
	\shorthandoff{=}  
	\includegraphics[scale=0.72]{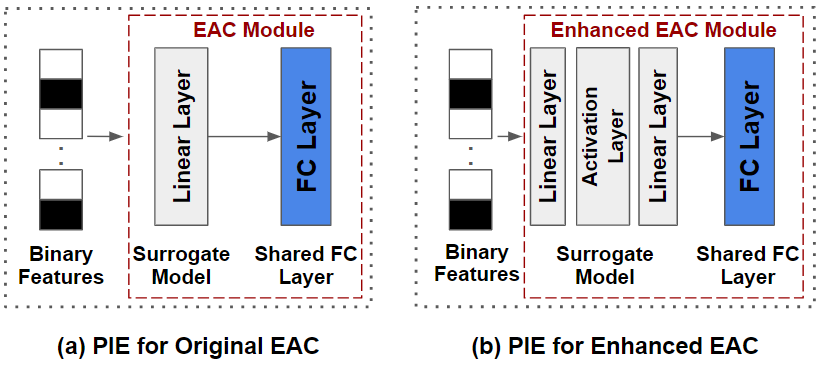}
	\shorthandon{=} 
	\caption{(a) shows the PIE for the original explain-any-concept (EAC) as described in \cite{eac}. (b) shows our proposed PIE which includes a non-linear surrogate model.}
	\label{enh}
\end{figure*}

Our approach builds upon the recently published "Explain Any Concept: Segment Anything Meets Concept-Based Explanation" paper and we, essentially, focus on improving the performance of the surrogate model as described in the EAC framework. The EAC framework\footnote{\url{https://github.com/Jerry00917/samshap}} leverages SAM for automatic visual concept extraction, a Per-Input Equivalence (PIE) surrogate model for efficient target behavior approximation, and utilizes Shapley values to quantify concept contributions to model predictions. In particular, the EAC framework uses only one trainable linear layer for the surrogate model. In the original surrogate architecture, the authors freeze the shared FC layer parameters when training the surrogate model. As such, only one linear layer is being trained. However, a trainable nonlinear network, by being more flexible in terms of its mathematical modelling capacity, can help improve the performance of the EAC framework. Therefore, we introduce using a trainable nonlinear network within the surrogate model in this paper. To keep the computational burdens minimal, we add only one {\it additional} linear layer with an activation function. That way, we introduce nonlinearity, within the surrogate model, increasing its modeling capability where two consecutive layers are trained while freezing the shared FC layer parameters. 
The overview of our enhanced EAC approach with nonlinear surrogate model is shown in Figure~\ref{dem}.  In the framework, the input is the image and the output is the explained image where the highlighted area shows the most important part used in the network for decision making.
The input image is first fed into a segmentation model (such as SAM \cite{sam}) and the corresponding concepts are obtained as the output of that segmentation model. Next those concepts, are first hot encoded and then fed into the surrogate model. The output of the surrogate model is the softmax output with confidence scores. Those confidence scores are given into an Monte-Carlo sampling process to estimate the final Shapley values for each concept. Figure \ref{enh} shows the original linear surrogate model in (a) as used in the EAC framework \cite{eac} and our introduced nonlinear surrogate model in (b).

The Enhanced EAC framework, similar to the original EAC framework, contains three phases and our modifications are done within the phase two. We kept phase one and three the same as in the original EAC framework \cite{eac}.

\textbf{Phase One: Concept Discovery:}
In this first phase, the objective is to delineate an input image into a collection of visual concepts that are semantically meaningful. This is achieved by employing the SAM model, which is, currently, a state of the art instance segmentation algorithm. Unlike preceding methods that utilized superpixel techniques, an instance segmentation based technique better ensures that the resulting concepts are not only derived from the image's features but are also meaningful to human observers. For a given image \( x \), SAM yields a set \( \ C = \{c_1, c_2, \ldots, c_n\} \), where \( n \) is the total number of concepts. Obtaining those $n$ meaningful concepts (as image masks) is the main goal in this phase. 

\textbf{Phase Two: Per-Input Equivalence (PIE):}
In this second phase, the PIE method is used which introduces a surrogate model designed to emulate the target network's behaviour on individual inputs, approximating the computation of Shapley values. The surrogate model is trained and optimized by using the cross-entropy loss, incorporating a shared FC layer (with frozen weights) from the target model to facilitate accurate predictions. This is formally defined as:
\[ f'(b) := f_{FC}(\mathbf{h}(b)) \]
where \( f' \) is the PIE model, \( b \) is the one-hot encoding of $n$ concepts (binary features), \( h \) is the surrogate model which mimics the feature extractor of the target model, and \( f_{FC} \) is the fully-connected (FC) layer of the original target model.\\
The training data for training the surrogate model is obtained by sampling the concepts in \( C \) and the corresponding probability distribution of the original target model by masking the concepts in \( C \). Then, the surrogate model is trained by plugging the FC layer (\( f_{FC} \)) of the target model (see Figure \ref{dem}) into \( f' \) and optimize \( h \) with the cross-entropy loss, where the weights of \( f_{FC} \) are frozen. Given that the surrogate model is much smaller than the target model, the PIE scheme can significantly reduce the cost of computing the Shapley values.

In this paper, we introduce using a non-linear model to represent \( h \). In the original EAC framework, the surrogate model, \( h \), is represented by training only one FC layer. In our approach, we use one FC layer followed by an activation function (such as tanh, ReLU, or sigmoid \cite{tanhsigrelu}), followed by another FC layer to represent the surrogate model,\( h \), therefore, in total, we have two trainable FC layers in the surrogate model. Figure \ref{enh} demonstrates the PIE architectures of the original EAC framework and our enhanced EAC framework. 

\textbf{Phase Three: Concept-based Explanation:}
Once the set of visual concepts \( C \) is identified as described in the first phase, this phase (phase three) aims to predict the final Shapley values of a given input image for explainability.  The Shapley values are estimated by applying a Monte Carlo sampling process on the outputs of the PIE surrogate model. The explanation $E$ is a subset of $C$, i.e., \( E \subseteq C \). The marginal contribution of the $i^{th}$ concept \( c_i \) can be represented by the difference between the model's prediction on the set \( S \setminus \{c_i\} \) and \( S \), where \( S \subseteq C \setminus \{c_i\} \). As such,  the individual contribution of \( c_i \) is defined as \cite{eac}:
\[
\Delta c_i(S) = u(S \cup \{c_i\}) - u(S)
\]
where \( u(S) \) represents the prediction of the target model based on only the concepts that are in \(S\) (the concepts that are not in \(S\) are masked). Since the surrogate model aproximates the original (target) model, we can re-write the above-given equation in terms of the surrogate model \( f' \). The Shapley value of the concept \( c_i \) is then defined as \cite{eac}:
\[
\phi_{c_i}(x) = \frac{1}{n} \sum_{k=1}^{n} \frac{1}{\binom{n-1}{k-1}} \sum_{S \subseteq S_{k}(i)} \Delta c_i(S)
\]
where \( S_k(i) \) is the set of all the coalitions that does not contain \( c_i \), with size \( k \). Since the size of all coalitions is prohibitively large, the EAC framework approximates the Shapley value by using Monte Carlo sampling. In particular, the framework samples \( K \) coalitions for each concept and approximates the Shapley value as \( \hat{\phi}_{c_i}(x) = \frac{1}{K} \sum_{k=1}^{K} \Delta_{c_i}(S_k) \), where \( S_k \) is the \( k^{th} \) sampled coalition. Selecting the subset of concepts maximizing the Shapley value determines the optimal explanation:
\[
E = \arg\max_{E \subseteq C} \phi_{E}(x)
\]
where \(\phi_E(x) = \sum_{c_{i} \in E} \hat{\phi}_{c_i}(x)\) is the  Shapley value of \(E(x)\). Finally, the masked image (where the concepts in \(C \setminus E\) are masked) represents the visual explainability for the user (further details can be found in \cite{eac}).

\section{Experiments}
Our experiments are conducted on two different datasets: ImageNet \cite{imagenet} and COCO \cite{coco}. {\it Faithfulness}, {\it Efficiency} and {\it Understandability} are important measures to consider for explainability. In this paper, we use area under the curve (AUC) value for measuring faithfulness, and per-input equivalence (PIE) time for measuring efficiency as described in \cite{eac}.  
In our experiments, since we introduced an improvement on the EAC model, we mainly compare our proposed solution (EEAC) to the original EAC model. In all of our experiments, all the models are trained for 50 epochs, individually; the learning rate was set to 0.001 and Adam optimizer was used. We tested the performance of EEAC by using different activation functions including tanh, sigmoid and ReLU. 

We assess the performance of both of the original and enhanced EAC models on ImageNet and COCO datasets, where we randomly sample 10,000 images from each dataset to explain. Both models employ the ResNet-50 model \cite{resnet} pre-trained on these datasets as the target DNN. Our evaluation metrics are based on insertion and deletion schemes, as also used in \cite{eac}. Insertion and deletion experiments \cite{insdelt} are leveraged to measure the faithfulness of the EAC explanations. These involve first altering the model's input by progressively adding (insertion) or removing (deletion) concept features from the most important to the least important ones; then computing the Area Under the Curve (AUC) value for each model. Higher AUC value for insertion, and lower AUC value for deletion indicate greater faithfulness. We run our experiments for insertion and deletion separately three times and report their average values in Table \ref{tab:imagenet} and Table \ref{tab:coco} where the best results are shown in bold.

\begin{table}[!b]
\centering
\resizebox{.485\textwidth}{!}{%
\begin{tabular}{|l|c|c|c|c|}
\hline
 & \multicolumn{2}{c|}{ImageNet / \textbf{Insertion}} & \multicolumn{2}{c|}{ImageNet / \textbf{Deletion}} \\
\cline{2-5}
\textbf{Model} & \textbf{AUC} & \textbf{PIE time (sec.)} & \textbf{AUC} & \textbf{PIE time (sec.)} \\
\hline
EEAC with Tanh   & \textbf{85.60} & 255 & \textbf{22.10} & 255 \\
EEAC with Sigmoid   & 83.57 & 255 & 24.43 & 255 \\
EEAC with ReLU   & 85.20 & 255 & 22.63 & 255 \\
Original EAC   & 84.04 & \textbf{250} & 23.56  & \textbf{250} \\
\hline
\end{tabular}
}
\caption{Experimental results on the ImageNet dataset for insertion and deletion. The results are shown for Enhanced EAC (ours) and the original EAC model \cite{eac} in terms of both area under the curve (AUC) and PIE time (in seconds).}
\label{tab:imagenet}
\end{table}

\begin{table}[!t]
\centering
\resizebox{.485\textwidth}{!}{%
\begin{tabular}{|l|c|c|c|c|}
\hline
 & \multicolumn{2}{c|}{COCO / \textbf{Insertion}} & \multicolumn{2}{c|}{COCO / \textbf{Deletion}} \\
\cline{2-5}
\textbf{Model} & \textbf{AUC} & \textbf{PIE time (sec.)} & \textbf{AUC} & \textbf{PIE time (sec.) } \\
\hline
EEAC with Tanh   & \textbf{85.22} & 217 & \textbf{16.11} & 217 \\
EEAC with Sigmoid    & 83.61 & 217 & 17.82 & 217 \\
EEAC with ReLU    & 84.90 & 217 & 16.44 & 217 \\
Original EAC   & 83.87 & \textbf{212} & 17.31  & \textbf{212} \\
\hline
\end{tabular}
}
\caption{Experimental results on the COCO dataset for insertion and deletion. The results are shown for Enhanced EAC (ours) and the original EAC model \cite{eac} in terms of both area under the curve (AUC) and PIE time (in seconds).}
\label{tab:coco}
\end{table}
As our preliminary results indicate, the introduction of an additional linear layer coupled with an activation layer, which introduces non-linearity, has provided a more nuanced approximation of the target DNN. This complexity in the surrogate model has translated into more faithful explanations, as evidenced by the improved AUC scores in Table \ref{tab:imagenet} and Table \ref{tab:coco}. {\it Tanh} yielded the best results among the tested activation functions. Notably, this increase in model fidelity has not come at the cost of efficiency. However, computational overhead due to the additional layers is minimal, preserving the model's practical applicability (as shown in Table I and II). Our preliminary results affirm the hypothesis that using a nonlinear model for the surrogate model can better capture the target model's behavior, thereby introducing room for more accurate explanations.

\section{Conclusion}
In this paper, we studied the performance of using and training a non-linear surrogate model within the recently proposed explain any concept (EAC) framework where the original EAC framework trained only one linear layer within the surrogate model. Our preliminary results indicate that training a nonlinear surrogate model within the EAC framework yields more accurate explainability for classification. Our proposed trainable nonlinear surrogate model has resulted in explanations that are more faithful to the target DNN. Nonlinearity can be obtained in various forms, and future work may include studying the performance of different nonlinear models within the surrogate model.
%

{\small
\bibliographystyle{ieeetr}
\bibliography{egbib.bib}

\begin{thebibliography}{10}

\bibitem{b1}
W.~Rawat and Z.~Wang, ``Deep convolutional neural networks for image
  classification: A comprehensive review,'' {\em Neural computation}, vol.~29,
  no.~9, pp.~2352--2449, 2017.

\bibitem{sahin2022yolodrone}
O.~Sahin and S.~Ozer, ``Yolodrone+: Improved yolo architecture for object
  detection in uav images,'' in {\em 2022 30th Signal Processing and
  Communications Applications Conference (SIU)}, pp.~1--4, IEEE, 2022.

\bibitem{b2}
C.~Szegedy, A.~Toshev, and D.~Erhan, ``Deep neural networks for object
  detection,'' {\em Advances in neural information processing systems},
  vol.~26, 2013.

\bibitem{valiente2020connected}
R.~Valiente, M.~Zaman, Y.~P. Fallah, and S.~Ozer, ``Connected and autonomous
  vehicles in the deep learning era: A case study on computer-guided
  steering,'' in {\em Handbook Of Pattern Recognition And Computer Vision},
  pp.~365--384, World Scientific, 2020.

\bibitem{OZER2}
H.~E. Ilhan, S.~Ozer, G.~K. Kurt, and H.~A. Cirpan, ``Offloading deep learning
  empowered image segmentation from uav to edge server,'' in {\em 2021 44th
  International Conference on Telecommunications and Signal Processing (TSP)},
  pp.~296--300, IEEE, 2021.

\bibitem{OZER}
S.~{\"O}zer, M.~Ege, and M.~A. {\"O}zkanoglu, ``Siamesefuse: A computationally
  efficient and a not-so-deep network to fuse visible and infrared images,''
  {\em Pattern Recognition}, vol.~129, p.~108712, 2022.

\bibitem{b4}
F.~Xu, H.~Uszkoreit, Y.~Du, W.~Fan, D.~Zhao, and J.~Zhu, ``Explainable ai: A
  brief survey on history, research areas, approaches and challenges,'' in {\em
  Natural Language Processing and Chinese Computing: 8th CCF International
  Conference, NLPCC 2019, Dunhuang, China, October 9--14, 2019, Proceedings,
  Part II 8}, pp.~563--574, Springer, 2019.

\bibitem{b5}
J.~Li, K.~Kuang, L.~Li, L.~Chen, S.~Zhang, J.~Shao, and J.~Xiao,
  ``Instance-wise or class-wise? a tale of neighbor shapley for concept-based
  explanation,'' in {\em Proceedings of the 29th ACM International Conference
  on Multimedia}, pp.~3664--3672, 2021.

\bibitem{b6}
J.~H.-w. Hsiao, H.~H.~T. Ngai, L.~Qiu, Y.~Yang, and C.~C. Cao, ``Roadmap of
  designing cognitive metrics for explainable artificial intelligence (xai),''
  {\em arXiv preprint arXiv:2108.01737}, 2021.

\bibitem{b7}
L.-V. Herm, ``Impact of explainable ai on cognitive load: Insights from an
  empirical study,'' {\em arXiv preprint arXiv:2304.08861}, 2023.

\bibitem{b8}
R.~C. Fong and A.~Vedaldi, ``Interpretable explanations of black boxes by
  meaningful perturbation,'' in {\em Proceedings of the IEEE international
  conference on computer vision}, pp.~3429--3437, 2017.

\bibitem{b9}
B.~Zhou, A.~Khosla, A.~Lapedriza, A.~Oliva, and A.~Torralba, ``Learning deep
  features for discriminative localization,'' in {\em IEEE conference on
  computer vision and pattern recognition}, pp.~2921--2929, 2016.

\bibitem{sam}
A.~Kirillov, E.~Mintun, N.~Ravi, H.~Mao, C.~Rolland, L.~Gustafson, T.~Xiao,
  S.~Whitehead, A.~C. Berg, W.-Y. Lo, {\em et~al.}, ``Segment anything,'' {\em
  arXiv preprint arXiv:2304.02643}, 2023.

\bibitem{eac}
A.~Sun, P.~Ma, Y.~Yuan, and S.~Wang, ``Explain any concept: Segment anything
  meets concept-based explanation,'' {\em arXiv preprint arXiv:2305.10289},
  2023.

\bibitem{shap}
L.~S. Shapley {\em et~al.}, ``A value for n-person games,'' 1953.

\bibitem{mc}
E.~Song, B.~L. Nelson, and J.~Staum, ``Shapley effects for global sensitivity
  analysis: Theory and computation,'' {\em SIAM/ASA Journal on Uncertainty
  Quantification}, vol.~4, no.~1, pp.~1060--1083, 2016.

\bibitem{modelspec1}
S.~M. Lundberg, G.~Erion, H.~Chen, A.~DeGrave, J.~M. Prutkin, B.~Nair, R.~Katz,
  J.~Himmelfarb, N.~Bansal, and S.-I. Lee, ``From local explanations to global
  understanding with explainable ai for trees,'' {\em Nature machine
  intelligence}, vol.~2, no.~1, pp.~56--67, 2020.

\bibitem{modelspec2}
M.~Ancona, C.~Oztireli, and M.~Gross, ``Explaining deep neural networks with a
  polynomial time algorithm for shapley value approximation,'' in {\em
  International Conference on Machine Learning}, pp.~272--281, PMLR, 2019.

\bibitem{surrogate1}
S.~M. Lundberg and S.-I. Lee, ``A unified approach to interpreting model
  predictions,'' {\em Advances in neural information processing systems},
  vol.~30, 2017.

\bibitem{surrogate2}
I.~Covert and S.-I. Lee, ``Improving kernelshap: Practical shapley value
  estimation using linear regression,'' in {\em International Conference on
  Artificial Intelligence and Statistics}, pp.~3457--3465, PMLR, 2021.

\bibitem{silency}
K.~Simonyan, A.~Vedaldi, and A.~Zisserman, ``Deep inside convolutional
  networks: Visualising image classification models and saliency maps,'' {\em
  arXiv preprint arXiv:1312.6034}, 2013.

\bibitem{deeplift}
A.~Shrikumar, P.~Greenside, and A.~Kundaje, ``Learning important features
  through propagating activation differences,'' in {\em International
  conference on machine learning}, pp.~3145--3153, PMLR, 2017.

\bibitem{lime}
M.~T. Ribeiro, S.~Singh, and C.~Guestrin, ``" why should i trust you?"
  explaining the predictions of any classifier,'' in {\em Proceedings of the
  22nd ACM SIGKDD international conference on knowledge discovery and data
  mining}, pp.~1135--1144, 2016.

\bibitem{shap2}
S.~M. Lundberg and S.-I. Lee, ``A unified approach to interpreting model
  predictions,'' {\em Advances in neural information processing systems},
  vol.~30, 2017.

\bibitem{cav}
B.~Kim, M.~Wattenberg, J.~Gilmer, C.~Cai, J.~Wexler, F.~Viegas, {\em et~al.},
  ``Interpretability beyond feature attribution: Quantitative testing with
  concept activation vectors (tcav),'' in {\em International conference on
  machine learning}, pp.~2668--2677, PMLR, 2018.

\bibitem{tanhsigrelu}
S.~R. Dubey, S.~K. Singh, and B.~B. Chaudhuri, ``Activation functions in deep
  learning: A comprehensive survey and benchmark,'' {\em Neurocomputing}, 2022.

\bibitem{imagenet}
J.~Deng, W.~Dong, R.~Socher, L.-J. Li, K.~Li, and L.~Fei-Fei, ``Imagenet: A
  large-scale hierarchical image database,'' in {\em 2009 IEEE conference on
  computer vision and pattern recognition}, pp.~248--255, Ieee, 2009.

\bibitem{coco}
T.-Y. Lin, M.~Maire, S.~Belongie, J.~Hays, P.~Perona, D.~Ramanan,
  P.~Doll{\'a}r, and C.~L. Zitnick, ``Microsoft coco: Common objects in
  context,'' in {\em Computer Vision--ECCV 2014: 13th European Conference,
  Zurich, Switzerland, September 6-12, 2014, Proceedings, Part V 13},
  pp.~740--755, Springer, 2014.

\bibitem{resnet}
K.~He, X.~Zhang, S.~Ren, and J.~Sun, ``Deep residual learning for image
  recognition,'' in {\em Proceedings of the IEEE conference on computer vision
  and pattern recognition}, pp.~770--778, 2016.

\bibitem{insdelt}
V.~Petsiuk, A.~Das, and K.~Saenko, ``Rise: Randomized input sampling for
  explanation of black-box models,'' {\em arXiv preprint arXiv:1806.07421},
  2018.

\end{thebibliography}
}

\end{document}